\documentclass[10pt,twocolumn,letterpaper]{article}

\usepackage{cvpr}
\usepackage{times}
\usepackage{epsfig}
\usepackage{graphicx}
\usepackage{amsmath}
\usepackage{amssymb}
\usepackage{booktabs}
\usepackage{multirow}
\usepackage{array}
\usepackage{verbatim}
\usepackage{subcaption}
\usepackage{adjustbox}
\usepackage{caption}
\usepackage{soul}
\usepackage[dvipsnames]{xcolor}
\usepackage{appendix}

\colorlet{Mycolor1}{green!10!orange!90!}

\usepackage[pagebackref=true,breaklinks=true,letterpaper=true,colorlinks,bookmarks=false]{hyperref}

\newcommand{\gallery}{gallery set}
\newcommand{\query}{query set}
\newcommand{\embtraining}{embedding training set}

\cvprfinalcopy 

\def\cvprPaperID{2166} 

\ifcvprfinal\fi
\begin{document}

\title{
Towards Backward-Compatible Representation Learning
}

\author{
Yantao Shen\footnotemark[1] \quad \ 
Yuanjun Xiong \quad \ 
Wei Xia \quad \ 
Stefano Soatto \\
AWS/Amazon AI\\
{\tt\small  ytshen@link.cuhk.edu.hk}, \quad
{\tt\small \{yuanjx, wxia, soattos\}@amazon.com}
}

\maketitle

\begin{abstract}
\footnotetext[1]{Currently at The Chinese University of Hong Kong. Work conducted while at AWS.}

We propose a way to learn visual features that are compatible with previously computed ones even when they have different dimensions and are learned via different neural network architectures and loss functions. Compatible means that, if such features are used to compare images, then ``new'' features can be compared directly to ``old'' features, so they can be used interchangeably. This enables visual search systems to bypass computing new features for all previously seen images when updating the embedding models, a process known as backfilling. Backward compatibility is critical to quickly deploy new embedding models that leverage ever-growing large-scale training datasets and improvements in deep learning architectures and training methods. We propose a framework to train embedding models, called backward-compatible training (BCT), as a first step towards backward compatible representation learning. In experiments on learning embeddings for face recognition, models trained with BCT successfully achieve backward compatibility without sacrificing accuracy, thus enabling backfill-free model updates of visual embeddings.
\end{abstract}

\section{Introduction}

Visual classification in an ``open universe'' setting is often accomplished by mapping each image onto a vector space using a function (``model'') implemented by a deep neural network (DNN). The output of such a function in response to an image is often called its ``embedding''~\cite{he2016deep,szegedy2015going}.
Dissimilarity between a pair of images can then be measured by some type of distance between their embedding vectors. A good embedding is expected to  cluster images belonging to the same class in the embedding space. 

\begin{figure}[h!]
\begin{tabular}{c}
  \includegraphics[scale=0.37]{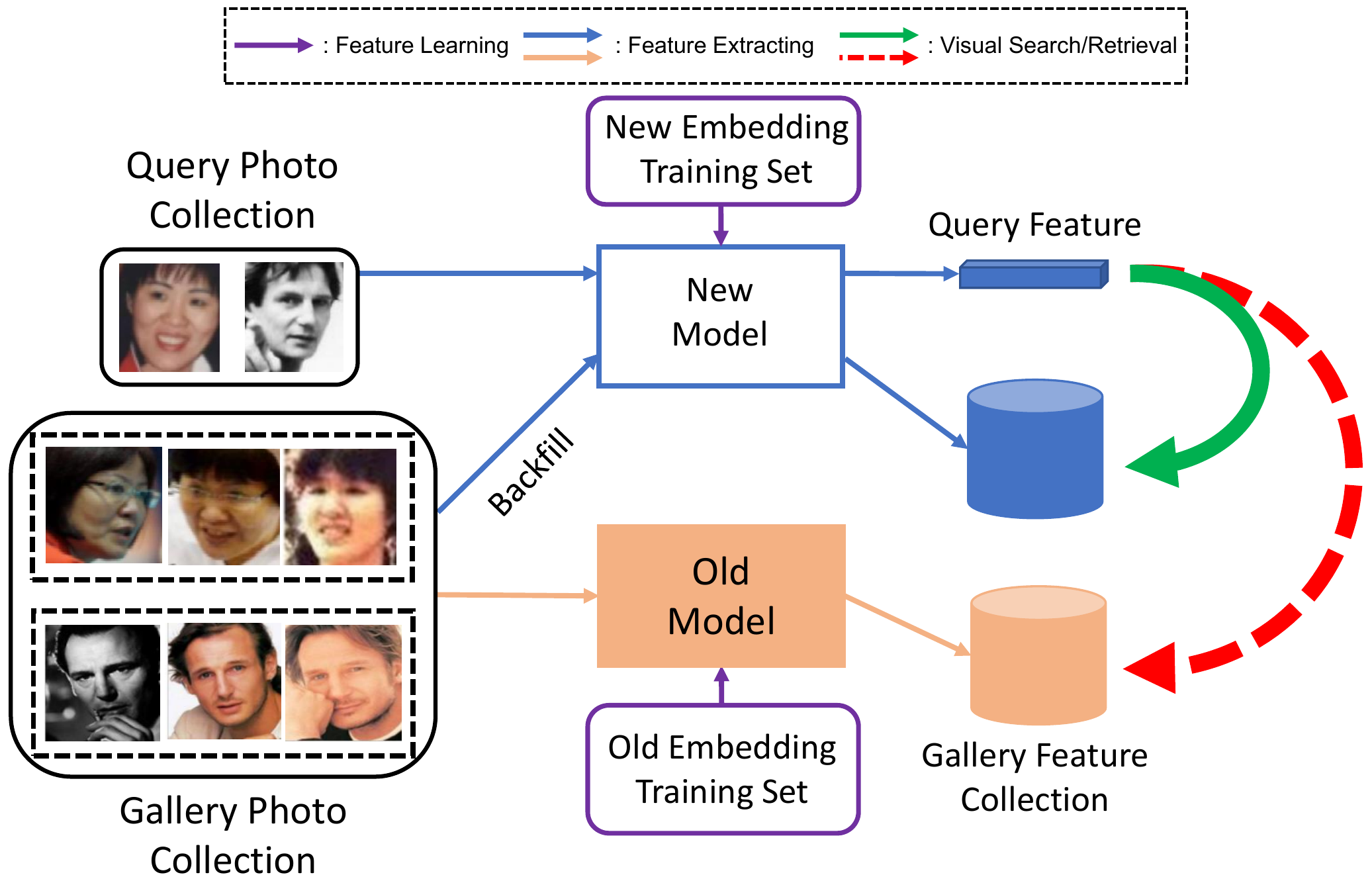}\\
  (a) Model update without backward compatible representation.\\
  \includegraphics[scale=0.37]{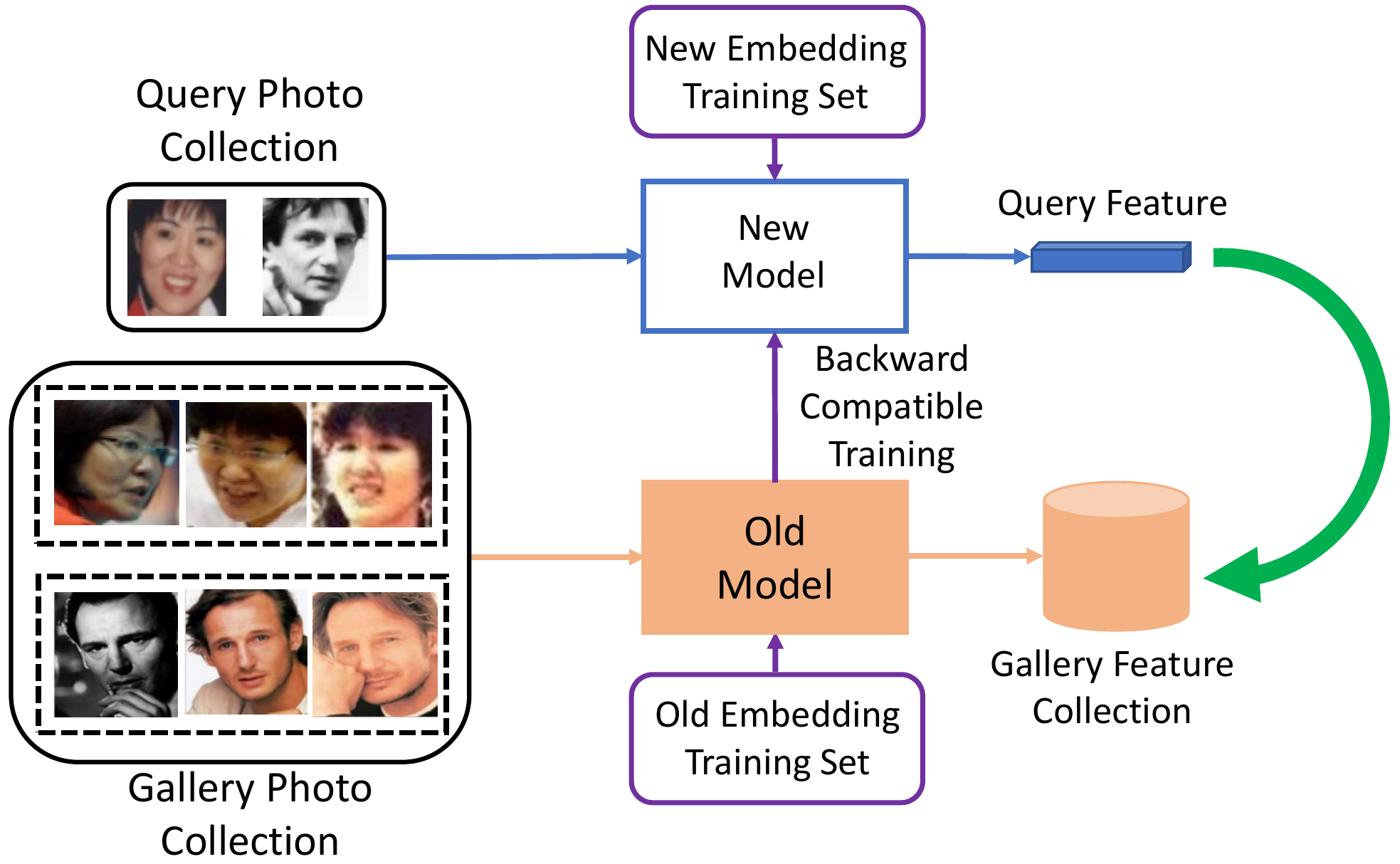}\\
  (b) Model update with backward compatible representation.
\end{tabular}
   \caption{
   Without backward compatible representation, to update the embedding model for a retrieval/search system, all previously processed gallery features have to be recomputed by the new model (backfilling), as the new embedding cannot be directly compared with the old one.
   With a backward compatible representation, direct comparison becomes possible, thus eliminating the need to backfill. 
   }
\label{fig:teaser_bct}
\end{figure}

As images of a new class become available, their embedding vectors are used to spawn a new cluster in the open universe, possibly modifying its metric to avoid crowding, in a form of ``life-long learning.'' This process is known as \emph{indexing}.
It is common in modern applications to have millions, in some cases billions, of images indexed into hundreds of thousands to millions of clusters.
This collection of images is usually referred to as the \emph{\gallery}. 
A common use for the indexed \gallery~ is to identify the closest clusters to one or a set of input images, a process known as \emph{visual search} or \emph{visual retrieval}. The set of input images for this task is known as the \emph{\query}.
Besides the gallery and the query set, there is usually a separate large repository of images used for training the embedding model~\cite{wang2018devil,Sun_2014_CVPR}, called the \emph{\embtraining}.

As time goes by, the datasets grow and the quality of the embeddings improves with newly trained  models
\cite{wen2016centerloss,wang2017normface,deng2019arcface,wang2018cosface}.
However, to harvest the benefits of new models, one has to use the new models to {\em re-process all images} in the \gallery~to generate their embedding and re-create the clusters, a process known as ``backfilling'' or ``re-indexing.''\footnote{The reader may have experienced this process when upgrading photo collection software, whereby the search feature is unavailable until the software has re-indexed the entire collection. 
This is a minor inconvenience in a personal photo collection, but for large-scale galleries, the cost of time and computation may be prohibitive, thus hampering the continuous update potential of the system.}
In this paper, we aim to design a system that enables new models to be deployed without having to re-index existing image collections. We call such a system {\em backfill-free}, the resulting embedding {\em backward-compatible representation}, and the enabling process {\em backward-compatible training} (BCT). 

We summarize our contributions as follow: 1) We formalize the problem of {\em backward compatible representation learning} in the context of open-set classification, or visual retrieval. The goal is to enable new models to be deployed without having to re-process the previously indexed \gallery. The core of this problem is backward compatibility, which requires a new embedding's output to be usable for comparisons against the old embedding model without compromising recognition accuracy. 2) We propose a novel backward compatible training (BCT) approach by adding an influence loss, which uses the learned classifier of the old embedding model in training the new embedding model. 3) We achieve backward compatible representation learning with minimal loss of accuracy, enabling backfill-free updates of the models. We empirically verify that BCT is robust against multiple changing factors in training the embedding models, \emph{e.g.}, neural network architectures, loss functions, and data growth. Finally, 4) we show that compatibility between multiple models can be attained via chain-like pairwise BCT training.

\subsection{Related Work}
\textbf{Embedding learning and open set recognition.}
Open-set visual recognition \cite{scheirer2012toward, scheirer2014probability} is relevant to retrieval~\cite{gordo2016deep, bendale2016towards}, face recognition~\cite{sun2014deep, wang2018devil, Cao18} and person re-identification~\cite{li2014deepreid,zheng2015scalable,ahmed2015improved}. Common approaches  involve extracting visual features to instantiate test-time classifiers~\cite{Sulc_2019_ICCV}. 
Deep neural networks (DNNs) are widely applied to 1) learn embedding models using closed-world classification as a surrogate task~\cite{krizhevsky2012imagenet,sun2014deep}, using various forms of loss functions~\cite{wang2018cosface,wang2017normface, xiao2016learning} and supervision methods~\cite{hinton2015distilling,krizhevsky2012imagenet}  to improve generalization 2) perform metric learning~\cite{parameswaran2010large} 
enforcing affinity for pairs~\cite{schroff2015facenet, ahmed2015improved} or triplets~\cite{parkhi2015deep, hermans2017defense} of representations in the embedding space.
Specifically, \cite{parameswaran2010large} learns a single metric that is compatible for all tasks in a multi-task learning setting.
Supervising representation learning with classifier weights from other versions of models was proposed in~\cite{wu2018unsupervised} for the task of unsupervised representation learning. 

\textbf{Learning across domains and tasks.}
In domain adaptation~\cite{panareda2017open,bousmalis2016domain,wang2018deep}, techniques such as MMD~\cite{yan2017mind} and related methods~\cite{tzeng2014deep,long2015learning, yan2017mind}, can be used to align the (marginal) distribution of the new and old classes, included those trained adversarially~\cite{hong2018conditional}.
Knowledge distillation in~\cite{hinton2015distilling,li2017learning} trains new models to learn from existing models, but, unlike backward compatible representation learning, knowledge distillation does not require the embedding of the new model and the existing one to be compatible in inference.
Continual learning~\cite{parisi2019continual, li2017learning}, transfer learning~\cite{bansal2019case}, and life-long learning~\cite{rebuffi2017icarl, hou2019learning} all deal with the cases where an existing model evolves over time. In  \cite{li2017learning},  model distillation is used as a form of regularization when introducing new classes. In \cite{rebuffi2017icarl}, old class centers are used to regularize samples from the new classes. Hou \etal~\cite{hou2019learning} proposed a framework for learning a unified classifier in the incremental setting. In~\cite{bansal2019case}, the authors designed a re-training loss function. Methods addressing catastrophic forgetting~\cite{li2017learning} are most closely related to our work, as a common reason for forgetting is the changing of the visual embedding for the subsequent classifiers.
The problem we are addressing differs in that we aim to achieve backward compatibility between any pair of old model and new model. The new model is not required to be initialized by nor share a similar network architecture as the old model.

\textbf{Compatible representations.}
In~\cite{li2016convergent}, the authors discuss the possible mapping between feature vectors from multiple models trained on the same dataset; \cite{yu2018slimmable, yu2019network, yu2019universally} introduce a design where multiple models with different channel widths but the same architecture share common subsets of parameters and representation, which implicitly imposes compatibility among representations from different models. we propose an approach to solve the problem of backward-compatibility in deep learning, in the sense defined in the previous section. We focus on open-universe classification using metric discriminant functions.

\section{Methodology}
We first formulate the problem of backward compatible representation learning, then describe a backward compatible training approach and its implementations.

\subsection{Problem Formulation}\label{sec:formalization}
    
    As a prototypical application, we use the following case of a photo collection ${\cal D}$, serving the role of the \emph{gallery}.  ${\cal D}$ is grouped into a number of classes or identities $Y = \{y_1, \dots, y_{{}_{N}}\}$. 
    We have an embedding model $\phi_{\rm old}$ that maps each image $x$ onto an embedding vector $z = \phi_{\rm old}(x) \in {\mathbb R}^{K_{\rm old}}$ with $x \in {\cal D}$. 
    The embedding model $\phi_{\rm old}$ is trained on an \embtraining, ${\cal T}_{\rm old}$. The embedding of any image  produced by  ${\cal D}$ can then be assigned to a class 
    through some distance $d:{\mathbb R}^K \times {\mathbb R}^K \rightarrow \mathbb{R}_+$. 
    In the simplest case, dropping the subscript ``old,'', each class in $Y$ is associated with a ``prototype'' or cluster center $\phi_i, \ i \in Y$.
    The vector $\phi_i$ for the class $i$ can be obtained by a set function $\phi_i = S(\{\phi(x)\}_{y(x) = i})$, where $y(x)$ is the corresponding class label of an image $x \in {\cal D}$. Common choices of the set function $S$ include averaging and attention models~\cite{yang2017neural}.
    A test sample $x$ is assigned to the class $y = \arg\min_{i \in Y} d(\phi(x), \phi_i) \in Y$. 
    Later, a new model $\phi_{\rm new}$ with  ${K_{\rm new}}$-dimensional embedding vectors becomes available, for instance trained with additional data in the new \embtraining, ${\cal T}_{\rm new}$ (${\cal T}_{\rm new}$ can be a superset of ${\cal T}_{\rm old}$), or using a different architecture. The new embedding $\phi_{\rm new}$ is potentially living in a different embedding space and it is possible that $K_{\rm new} \neq K_{\rm old}$. 
    
    To harvest the benefit of the new embedding model $\phi_{\rm new}$, we wish to use $\phi_{\rm new}$ to process any new images that in the \gallery~, ${\cal D}$, as well as images for the \query. Since the \gallery~could get additional images and clusters, we denote it as ${\cal D}_{\rm new} = {\cal D} \cup \{x | y = N+1, ..., N_{\rm new}\}$, where $N_{\rm new} $ is the number of clusters in ${\cal D}_{\rm new}$.
    Then, the question becomes how to deal with images in ${\cal D}$.
    In order to make the system backfill-free, we wish to directly use the already computed embedding from $\phi_{\rm old}$ for these images and obtain $\{\phi_i\}_{i \leq N}$ .
    Our goal, then, is to design a training process for the new embedding model $\phi_{\rm new}$ so that any test images can be assigned to classes, new or old, in ${\cal D}_{\rm new}$, without the need to compute $\phi_{\rm new}({\cal D})$, {\em i.e.}, to backfill. 
    The resulting embedding $\phi_{\rm new}$, is then backward-compatible with $\phi_{\rm old}$.
    
\subsection{Criterion for Backward Compatibility}

In a strict sense, a model $\phi_{\rm new}$ is backward compatible if
\begin{eqnarray}\label{eq:compatible_pairwise}
    d(\phi_{\rm new}(x_i), \phi_{\rm old}(x_j)) &\geq& d(\phi_{\rm old}(x_i), \phi_{\rm old}(x_j)), \nonumber \\
     & & \forall (i, j) \in \{(i, j) | y_i \neq y_j \}. \nonumber \\
    \mathrm{and}, & & \nonumber \\
    d(\phi_{\rm new}(x_i), \phi_{\rm old}(x_j)) &\leq& d(\phi_{\rm old}(x_i), \phi_{\rm old}(x_j)), \nonumber \\ 
    & & \forall (i, j) \in \{(i, j) | y_i = y_j \}.
\end{eqnarray}
where $d(\cdot,\cdot)$ is a distance in the embedding space. 
These constraints formalize the fact that the new embedding, when used to compare against the old embedding, must be at least as good as the old one in separating images from different classes and grouping those from the same classes. Note that the solution $\phi_{\rm new} = \phi_{\rm old}$ is backward compatible. This trivial solution is excluded if the architectures are different, which is usually the case when updating a model. Although, to simplify the discussion, we assume the embedding dimensions for the two models to be the same  ($K_{\rm new} = K_{\rm old}$), our method is more general and not bound by this assumption. 

The criterion introduced in Eq.~\ref{eq:compatible_pairwise} entails testing the gallery exhaustively, which is intractable at large scale and in the open-set setting.
    On the other hand, suppose we have an evaluation metric, $\mathop{M} (\phi_{q}, \phi_{g}; {\cal Q}, {\cal D},)$ on some testing protocols, \eg, true positive identification rates for face search, where  ${\cal Q}$ denotes the query set, ${\cal D}$ denotes the gallery set, and we use $\phi_{q}$ for extracting the \query~ feature and $\phi_{g}$ for the \gallery.
    Then, the \textbf{empirical compatibility criterion}, for the application can be defined as
    \begin{equation} \label{eq:compatible_set}
    \mathop{M} (\phi_{\rm new}, \phi_{\rm old}; {\cal Q}, {\cal D}) > \mathop{M} (\phi_{\rm old}, \phi_{\rm old}; {\cal Q}, {\cal D}).
    \end{equation}
    
This criterion can be interpreted as follows: In an open-set recognition task with a fixed query set and a fixed gallery set, when the accuracy using $\phi_{\rm new}$ for queries without backfilling gallery images surpasses that of using $\phi_{\rm old}$, we consider backward compatibility achieved and backfill-free update feasible. 
Note that simply setting $\phi_{\rm new}$ to $\phi_{\rm old}$ will not satisfy this criterion.

\subsection{Baseline and paragon}~\label{sec:baseline}
A naive approach to train the model $\phi_{\rm new}$ to be compatible with $\phi_{\rm old}$, assuming they have the same dimension,
is to minimize the $\ell^2$ distance between their embeddings computed on the same images. This is enforced for every image in ${\cal T}_{\rm old}$, which is used to train $\phi_{\rm old}$. This criterion can be framed as an additive regularizer $R$ for the empirical loss $L(\phi_{\rm new})$ when training the new embedding as
%
$$\phi_{\rm new} = \arg\min_{\phi} L(\phi, {\cal T}_{\rm new}) + \lambda R(\phi), \ {\rm where}$$
\begin{equation}\label{eq:bct_l2}
 R(\phi) = \sum_{\stackrel{x \in  \cal{T}_{\rm old}}{}}
  \frac{1}{2}\|\phi(x)
  - \phi_{\rm old}(x)\|^2_2.
\end{equation}

We label the solution of the problem above  $\phi_{\rm new-\ell^2}$. Note that $\phi_{\rm old}$ will be fixed during  training of $\phi_{{\rm new}-\ell^2}$.
As we show in Sect.~\ref{sec:exp_direct},   $\phi_{\rm new-\ell^2}$ 
does \textbf{not} satisfy Eq.~\eqref{eq:compatible_set} and it will not converge to $\phi_{\rm old}$, since the training set has been changed to ${\cal T}_{\rm new}$.  So, this naive approach cannot be used to obtain a backward compatible representation. 

On the other hand, performing the backfill on ${\cal D}$ with the model $\phi_{\rm new}$, trained without any regularization, can be taken as a paragon. Since the embedding for ${\cal D}$ is re-computed, we can fully enjoy the benefit of $\phi_{\rm new}$ 
albeit at the cost of reprocessing the gallery. This sets the upper bound of accuracy for the backfill-free update, and thus the upper bound of the update gain. 

\subsection{Backward Compatible Training}
\label{sec:bct_basic}
We now focus on backward compatible training for    classification using the cross-entropy loss.
Let $\Phi$  be a model parametrized by two disjoint sets of weights, $w_c$ and $w_{\phi}$. The first parametrizes the classifier ${\kappa}$, or the ``head'' of the model, whereas the second parametrizes the embedding $\phi$, so that $\Phi(x) = \kappa_{w_c}(\phi_{w_{\phi}}(x))$. Now, the cross-entropy loss can be written as
\begin{equation}
    L(w_c, w_\phi; {\cal T}) = \sum_{(x_i, y_i) \in {\cal T}} -\log \kappa_{w_c}(\phi_{w_\phi}(x_i))_{y_i}.
\end{equation}
Note that the classifier $\kappa_{w_c}$ can take many forms, from the simple SoftMax~\cite{sun2014deep,krizhevsky2012imagenet} to recently proposed alternatives~\cite{liu2017sphereface, wang2017normface, wang2018cosface}.
The old model $\phi_{\rm old}$ is thus obtained by solving
\begin{equation}
    w_{c \ {\rm old}}, w_{\phi \ {\rm old}} = \arg\min_{w} L(w_c, w_\phi; {\cal T}_{\rm old}).
\end{equation}
As for the new model $\phi_{\rm new}$ , while ordinary training would yield
\begin{equation}
    w_{c\ {\rm new}}, w_{\phi \ {\rm new}} = \arg\min_{w} L(w_c, w_\phi; {\cal T}_{\rm new}),
\end{equation}
to ensure backwards-compatibility, we add a second term to the loss that depends on the classifier of the old model:
\begin{equation}
    w_{c\ {\rm new}}, w_{\phi \ {\rm new}} = \arg\min_{w} L_{\rm BCT} (w_c, w_\phi; {\cal T}_{\rm new}, {\cal T}_{\rm BCT}),
\end{equation}
where
\begin{multline}\label{eq:bct_loss}
   L_{\rm BCT}(w_c, w_\phi; {\cal T}_{\rm new}, {\cal T}_{\rm BCT}) = L(w_c, w_\phi; {\cal T}_{\rm new}) + \\ + \lambda L(w_{c \ {\rm old}}, w_\phi; {\cal T}_{\rm BCT}).
\end{multline}
We call the second term ``influence loss'' since it biases the solution towards one that can use the old classifier. Note that $w_{c \ {\rm old}}$ in the influence loss will be fixed during training.
Here, ${\cal T}_{\rm BCT}$ is a design parameter, referring to the set of images we apply the influence loss to. 
It can be either ${\cal T}_{\rm old}$ or ${\cal T}_{\rm new}$. The approach of using ${\cal T}_{\rm new}$ as ${\cal T}_{\rm BCT}$ will be introduced in Sect.~\ref{sec:bct_dataset}.
Note that the classifiers $\kappa$ of the new and old models can be different.
We call this method backward compatible training, and the result backward compatible representation or embedding, which we evaluate empirically in the next section.

\subsection{Learning with Backward Compatible Training}
In the proposed backward compatible training framework, there are several design choices to make.

\textbf{Form of the classifier}
The classifiers $\kappa$ of the new and old models
$\kappa_{\rm new}$ and $\kappa_{\rm old}$ can be of the same form, for instance  Softmax, angular SoftMax classifier~\cite{liu2017sphereface}, or cosine margin~\cite{wang2018cosface}.
They can also be of different forms, which is common in the cases where better loss formulations are proposed and applied to training new embedding models. 

\textbf{Backward compatibility training dataset.}
\label{sec:bct_dataset}
The most straightforward choice for the dataset ${\cal T}_{\rm BCT}$, on which we apply the influence loss, is ${\cal T}_{\rm old}$, which was used to train the old embedding $\phi_{\rm old}$. The intuition is that, since the old model $\phi_{\rm old}$ is optimized together with its classifier $\kappa_{{\rm old}} $ on the original training set ${\cal T}_{\rm old}$, a new embedding model having a low influence loss will work with the old model's classifier and thus with the embedding vectors from $\phi_{\rm old}$.
The second choice of ${\cal T}_{\rm BCT}$ is ${\cal T}_{\rm new}$; this means that we not only compute the influence loss on the old training data for $\phi_{\rm old}$, but also on the new training data. However, this choice poses a challenge in the computation of the loss value  $L(w_c, w_\phi; {\cal T}_{\rm new})$
for the images in $\{x | x \in {\cal T}_{\rm new}, x \notin {\cal T}_{\rm old}\}$, due to the unknown classifier parameters for the classes. We propose two rules for computing the loss value for these images: \\
\textbf{Synthesized classifier weights}. For classes in ${\cal T}_{\rm new}$ which are not in the set of classes in ${\cal T}_{\rm old}$, we create their ``synthesized'' classifier weights by computing the average feature vector of $\phi_{\rm old}$ on the images in each class. This approach is inspired by open-set recognition using the class vector $\phi_i$t as described in Sect.~\ref{sec:formalization}. We use averaging as the set function in this case. The synthesized classifier weights for the new classes are concatenated with the existing $w_c$
to form the classifier parameters for the influence loss term.\\
\textbf{Knowledge distillation}. We penalize the KL-divergence of the classifier output probabilities between using $\phi_{\rm new}$ and $\phi_{\rm old}$ with existing classifier parameters $w_c$.
This removes the requirement to add new classes in ${\cal T}_{\rm new}$ to the classifiers corresponding to $\phi_{\rm old}$.

Backward compatible training is not restricted to certain neural network architecture or loss function. 
It only requires that both the old and new embedding models be trained with classification-based losses, which is common in open-set recognition problems~\cite{ wang2018cosface, li2014deepreid}. 
It also does not need modification of the architecture nor of the parameters of the old model $\phi_{\rm old}$.

\section{Experiments}

We assess the effectiveness of the proposed backward compatibility training in 
face recognition.
We start with several baselines,
then test the hypothesis that BCT leads to backward compatible representation learning on two face recognition tasks: face verification and face search. 
Finally, we demonstrate the potential of BCT by applying it to the cases of multi-factor model changes and showing it is able to construct multiple compatible models.

\subsection{Datasets and Face Recognition Metrics}
We use the IMDB-Face dataset~\cite{wang2018devil} for training face embedding models. The IMDB-Face dataset contains about 1.7M images of 59K celebrities. 
For the openset test, we use the widely adopted IJB-C face recognition benchmark dataset~\cite{maze2018iarpa}. It has around $130$k images from 3,531 identities.
The images in IJB-C contain both still images and video frames.
We adopt the two standard testing protocols for face recognition: 1:1 verification and 1:N search (open set). 
For 1:1 verification, a pair of templates (a template contains one or more face images from the same person) are presented and the algorithm is required to decide whether they belong to the same person or not.
The evaluation metrics for this protocol are true acceptance rate (TAR) at different false acceptance rates (FAR). 
We present the results of TAR at the FAR of $10^{-4}$.

For 1:N search, a set of templates is first indexed as the~\gallery. Then each template in the~\query~is used to search against the indexed templates. 
The quality metrics for this protocols are true positive identification rates (TPIR) at different false positive identification rates (FPIR).
We present the results of TPIR at $10^{-2}$ FPIR.

\subsection{Implementation details}

We use 8 NVIDIA Tesla V-100 GPUs in training the embedding models. The input size of embedding models is set to $112 \times 112$ pixels~\cite{wen2016centerloss}. We use face mis-alignment and color distortion for data augmentation. Weight decay is set to $5 \times 10^{-4}$ and standard stochastic gradient descent (SGD) is used to optimize the loss. The initial learning rate is set to $0.1$ and decreased to $0.01$, $0.001$, and $0.0001$ after 8, 12, and 14 epochs, respectively. The training stops after 16 epochs. The batchsize is set to $320$. Unless stated otherwise, we use ResNet-101~\cite{he2016deep} as the backbone, a linear transform after its global average pooling layer to emit 128-dimensional feature vectors , and Cosine Margin Loss~\cite{wang2018cosface} with margin=$0.4$ as the loss function in our experiments.

\subsection{Measuring Backward-Compatibility}~\label{sec:compatible_measure}
\vspace{-10pt}

Based on the accuracy on the individual tests on the face recognition dataset, we can  test whether a pair of  models satisfies the empirical backward compatibility criterion. For a pair of models $(\phi_{\rm new}, \phi_{\rm old})$, on each evaluation protocol we test whether they satisfy Eq.~\eqref{eq:compatible_set}. If so, we consider the new model \textbf{backward compatible} with the old model in the corresponding task. 
In testing using the IJB-C 1:N protocol~\cite{maze2018iarpa}, we use the new model $\phi_{\rm new}$ to extract embeddings for the \query~ and the old model $\phi_{\rm old}$ to compute embeddings for~\gallery. For the IJB-C 1:1 verification protocol~\cite{maze2018iarpa}, we use $\phi_{\rm new}$ to extract the embeddings for the first template in the pair and $\phi_{\rm old}$ for the second.

To evaluate relative improvement brought by the backfill-free update, we define the~\textbf{update gain} as
\begin{equation}\hspace{-10pt}\label{eq:update_gain}
\small
    {\cal G}(\phi_{\rm new}, \phi_{\rm old}; {\cal Q}, {\cal D}) =  \frac{\mathop{M} (\phi_{\rm new}, \phi_{\rm old};{\cal Q}, {\cal D}) - \mathop{M} (\phi_{\rm old}, \phi_{\rm old};{\cal Q}, {\cal D)}}{ \mathop{M} (\phi_{\rm new}^*, \phi_{\rm new}^*;{\cal Q}, {\cal D}) - \mathop{M} (\phi_{\rm old}, \phi_{\rm old};{\cal Q}, \cal D)}.
\end{equation}
Here, $\mathop{M}(\phi_{\rm new}^*, \phi_{\rm new}^*; {\cal Q}, {\cal D})$ stands for the best accuracy level 
we can achieve from any variants of the new model by backfilling. 
It indicates the proportional gain we can obtain the performing backfill-free update compared with the update which performs the backfill regardless of the cost and interruption of services.
Note that the update gain is only valid when Eq.~\eqref{eq:compatible_set} is satisfied.

\begin{table}[t]
\captionsetup{width=0.5\textwidth}
\small
\begin{subtable}{1.05\linewidth}\hspace{0pt}
\centering
\renewcommand\arraystretch{0.8}
\setlength{\tabcolsep}{4.0pt}
\begin{tabular}{lccc}
\toprule
New Model & Old Model & Data & Additional Loss\\
\midrule
$\phi_{\rm old}$  & - & 50\%  & -\\
$\phi_{\rm new}^*$  & - & 100\%     & -\\
$\phi_{{\rm new}-\ell^2}$   &$\phi_{\rm old}$  & 100\%     & $\ell^2$ distance $\phi_{\rm old}$\\
$\phi_{{\rm new-LwF}}$ &$\phi_{\rm old}$ &  50\%  & Learning w/o Forgetting\\
$\phi_{{\rm new}-\beta}$ &$\phi_{\rm old}$  & 100\% &  Influence loss on ${\cal T}_{\rm old}$ \\
$\phi_{{\rm new}-\beta-kd}$ & $\phi_{\rm old}$ & 100\% &  Influence loss on ${\cal T}_{\rm new}$ \\
$\phi_{{\rm new}-\beta-sys}$ & $\phi_{\rm old}$ & 100\% &  Influence loss on ${\cal T}_{\rm new}$\\
\bottomrule
\end{tabular}
\vspace{0pt}
\caption{Training setting for different backward-compatible (new) models. 
`$\phi_{old}$': the compatible target (old) model for all new models. 
`$\phi_{{\rm new}-\ell^2}$': the new model regularized with $\ell^2$ distance to $\phi_{\rm old}$ . 
`$\phi_{{\rm new-LwF}}$': the new model that adopts learning w/o forgetting~\cite{li2017learning} in training.
`$\phi_{{\rm new}-\beta}$': the new model trained with the proposed BCT.
`$\phi_{{\rm new}-\beta-kd}$': the new model trained with the proposed BCT and  knowledge distillation for new classes in the new embedding training dataset. 
`$\phi_{{\rm new}-\beta-sys}$': the new model trained with proposed BCT and the synthesised classifiers for new classes  in  the  new  embedding  training dataset. 
}
\label{tab:bct_baselines_setting}
\end{subtable}
\hfill
\begin{subtable}{1.0\linewidth}\hspace{0pt}
\setlength{\tabcolsep}{1.4pt}
\begin{tabular}{lccccc}
\toprule
\multirow{2}{*}{\shortstack{Comparison Pair}} 
&\multirow{2}{*}{\shortstack{Veri.\\Acc.}}
&\multirow{2}{*}{\shortstack{Backward\\Compatible?}}  &\multirow{2}{*}{\shortstack{Update\\Gain (\%)}} 
& \multirow{2}{*}{\shortstack{Absolute\\Gain}} \\
& \multicolumn{1}{c}{ }  \\

\midrule
$(\phi_{\rm old}, \phi_{\rm old}$) \begin{footnotesize}(Lower Bound)\end{footnotesize} &77.86 & - & - & - \\
\hline
$(\phi_{\rm new}^*, \phi_{\rm old})$&0.0 &  $\times$ & - & - \\
$(\phi_{{\rm new}-\ell^2}, \phi_{\rm old})$ &3.10 & $\times$ & - & -\\
$(\phi_{{\rm new-LwF}}, \phi_{\rm old})$ & 77.26 & $\times$ & -  & -\\
$(\phi_{{\rm new}-\beta}, \phi_{\rm old})$ \begin{footnotesize}(Ours)\end{footnotesize}&80.25 &$\surd$ & 26.26  &2.39 \\
$(\phi_{{\rm new}-\beta-kd}, \phi_{\rm old})$ \begin{footnotesize}(Ours)\end{footnotesize}&80.34 &$\surd$ & 27.25  & 2.48\\
$(\phi_{{\rm new}-\beta-sys}, \phi_{\rm old})$ \begin{footnotesize}(Ours)\end{footnotesize}&80.59 &$\surd$ & 30.00  & 2.73\\
\hline
$(\phi_{\rm new}^*, \phi_{\rm new}^*)$ \begin{footnotesize}(Upper Bound)\end{footnotesize} &86.96 & -& - & 9.1\\
\bottomrule
\end{tabular}
\caption
{
Experiments on the IJB-C 1:1 verification task. The verification accuracy evaluation metric is TAR (\%)@FAR=$10^{-4}$.
}
\label{tab:bct_baselines_result_1v1}
\end{subtable}
\begin{subtable}{1.0\linewidth}\hspace{0pt}
\setlength{\tabcolsep}{1.4pt}
\begin{tabular}{lccccc}
\toprule
\multirow{2}{*}{\shortstack{Comparison Pair}} 

&\multirow{2}{*}{\shortstack{Search\\Acc.}}
&\multirow{2}{*}{\shortstack{Backward\\Compatible?}} 
&\multirow{2}{*}{\shortstack{Update\\Gain (\%)}} 
&\multirow{2}{*}{\shortstack{Absolute\\Gain}}\\

& \multicolumn{1}{c}{ }  \\
\midrule
$(\phi_{\rm old}, \phi_{\rm old})$ \begin{footnotesize}(Lower Bound)\end{footnotesize}&59.34 & - &  - & -   \\
\hline
$(\phi_{\rm new}^*, \phi_{\rm old})$&0.0 & $\times$ & - & -\\
$(\phi_{{\rm new}-\ell^2}, \phi_{\rm old})$&0.50 & $\times$ &- & - \\
$(\phi_{{\rm new-LwF}}, \phi_{\rm old})$ &59.27 & $\times$ & - & -\\
$(\phi_{{\rm new}-\beta},\phi_{\rm old})$ \begin{footnotesize}(Ours)\end{footnotesize}&67.23 & $\surd$ & 44.98 & 7.89 \\
$(\phi_{{\rm new}-\beta-kd}, \phi_{\rm old})$ \begin{footnotesize}(Ours)\end{footnotesize}&69.02 &$\surd$ & 55.11  & 9.68 \\
$(\phi_{{\rm new}-\beta-sys}, \phi_{\rm old})$ \begin{footnotesize}(Ours)\end{footnotesize}&70.70 &$\surd$ & 64.77  & 11.36  \\
\hline
$(\phi_{\rm new}^*, \phi_{\rm new}^*)$ \begin{footnotesize}(Upper Bound)\end{footnotesize} &76.88 & - & - & 17.54 \\
\bottomrule
\end{tabular}
\caption{Experiments on the IJB-C 1:N search task. The search accuracy evaluation metric is TPIR(\%)@FPIR=$10^{-2}$.
}
\label{tab:bct_baselines_result_1vn}
\end{subtable}
\caption{Illustration of simple baselines and our proposed approach in backward compatibility test. In face search we use the first model of each comparison pair ]for the query set and the second for the gallery set, and in face verification for the first and second template, respectively.  The details of model training setting are illustrated in Tab.~\ref{tab:bct_baselines_setting}. We report the relative update gain defined in Eq.~\ref{eq:update_gain}.}
\vspace{-10pt}
\label{tab:bct_baselines}
\end{table}

\begin{table}{\hspace{-10pt}}
\small
\renewcommand\arraystretch{0.8}
\setlength{\tabcolsep}{1.0pt}
\begin{tabular}{lccccc}
\toprule
\multirow{2}{*}{\shortstack{Comparison Pair}}  & \multirow{2}{*}{\shortstack{IJB-C 1:1 Verifi.\\\begin{footnotesize}TAR (\%)@FAR=$10^{-4}$\end{footnotesize}}} & \multirow{2}{*}{\shortstack{IJB-C 1:N Retri.\\\begin{footnotesize}TPIR(\%)@FPIR=$10^{-2}$\end{footnotesize}}} \\
& \multicolumn{1}{c}{ }  \\
\midrule
$(\phi_{\rm old}, \phi_{\rm old})$ & 77.86 & 59.34 \\
$(\phi_{\rm new-\beta}, \phi_{\rm new-\beta})$ &85.36  &73.86\\
$(\phi_{{\rm new}-\beta-kd}, \phi_{{\rm new}-\beta-kd})$ &84.95  & 73.56\\
$(\phi_{{\rm new}-\beta-sys}, \phi_{{\rm new}-\beta-sys})$ &85.58  & 74.40\\
$(\phi_{\rm new}^*, \phi_{\rm new}^*)$ &86.96  &76.88 \\
\bottomrule
\end{tabular}
\vspace{-5pt}
\caption{Backward compatibility test for old model $\phi_{\rm old}$, BCT trained model $\phi_{\rm new-\beta}$, $\phi_{\rm new-\beta-kd}$, $\phi_{\rm new-\beta-sys}$ and paragon/upper bound model $\phi_{\rm new}^*$. The results shows that BCT does not lead to significant accuracy drop compared with paragon/upper bound model.
}
\vspace{-10pt}
\label{tab:bct_upper_bound_res}
\end{table}

\subsection{Baseline comparisons}\label{sec:bct_necessity}
The first hypothesis to be tested is whether BCT is necessary at all: is it possible to achieve backward compatibility with a more straightforward approach? In this section, we experiment with several  baseline approaches and validate the necessity of BCT.

\textbf{Independently trained $\phi_{\rm new}$ and $\phi_{\rm old}$}
\label{sec:exp_direct}
The first sanity-check is to directly compare the embedding of two versions of models trained independently. A similar experiment is done in~\cite{li2016convergent}~for multiple close-set classification models trained on the same dataset. 
Here we present two models. The $\phi_{\rm old}$ is trained with the randomly sampled $50\%$ IDs subset of the IMDBFace dataset~\cite{wang2018devil}. The new model is trained on the full IMDBFace dataset~\cite{wang2018devil}.
This emulates the case where a new embedding model becomes available when the sizes of embedding training dataset grows. 
We name the new model $\phi_{\rm new}^*$ according to the experiment in Sect.~\ref{sec:upper_bound}, showing that it currently achieves the best accuracy among all new models with the same setting.
We directly test the compatibility of this pair of models $(\phi_{\rm new}^*, \phi_{\rm old})$ following the procedure described in Sect.~\ref{sec:compatible_measure}. The results are illustrated in Tab.~\ref{tab:bct_baselines}.
In the backward test for both protocols, we observed almost $0\%$ accuracy. Unsurprisingly, 
independently trained $\phi_{\rm new}$ and $\phi_{\rm old}$ does not
naturally satisfy our compatibility criterion.

\textbf{Does the naive baseline with $\ell^2$-distance work?}
\label{sec:exp_l2}
In Sect.~\ref{sec:baseline} we described the naive approach of adding the $\ell^2$-distance between the new and old embeddings as a regularizer when training the new model. 
We train a new model using the same old model above and train the new model with the loss function \eqref{eq:bct_l2} on the whole IMDBFace dataset~\cite{wang2018devil}. 
We name this model $\phi_{{\rm new}-\ell^2}$ to reflect that it is $\ell^2$-distance regularized towards the old model.
The same backward compatibility test is conducted with this pair of models $(\phi_{{\rm new}-\ell^2}, \phi_{\rm old})$ on the same two protocols described in the previous baseline. The results are shown in Tab.~\ref{tab:bct_baselines}.
We can observe that this approach only leads to slightly above $0\%$ backward test accuracy, which means the new model $\phi_{{\rm new}-\ell^2}$ is far from satisfying the compatibility criterion.
One possible reason is that imposing an $\ell_2$ distance penalty creates a bias that is too local  and restrictive to allow the new model to satisfy the compatibility constraints.

\begin{table*}
\small
\renewcommand\arraystretch{0.9}
\setlength{\tabcolsep}{3pt}
\renewcommand\arraystretch{0.7}
\begin{subtable}{1.0\textwidth}\centering
\begin{tabular}{lcccccc}
\toprule
New Model  &Old Model & Training Data Usage & Feat. Dim. & Model Arc. & Classifier & Additional Loss \\
\midrule
$\phi_{\rm old}$ & -& 50\% & 128  & ResNet-101 & Cosine Margin &  - \\
$\phi_{{\rm new}-\beta}^{\rm R152} $ &$\phi_{\rm old}$& 100\%  & 128 & ResNet-152 & Cosine Margin   & Influence loss on  ${\cal T}_{\rm old}$   \\
$\phi_{{\rm new}-\beta}^{\rm R152+256D}$  &$\phi_{\rm old}$ & 100\%  & 256 & ResNet-152 & Cosine Margin   & Influence loss on  ${\cal T}_{\rm old}$   \\
$\phi_{{\rm new}-\beta}^{\rm ReLU}$  &$\phi_{\rm old}$ & 100\%  & 128 & ResNet-152 & Cosine Margin   &  Influence loss on  ${\cal T}_{\rm old}$    \\
\hline
$\phi_{\rm old}^{\rm NS}$  & - & 100\%  & 128 & ResNet-101 & Norm-Softmax  &  -   \\
$\phi_{{\rm new}-\beta}^{\rm Cos-NS}$& $\phi_{\rm old}^{\rm NS}$   & 100\%  & 128 & ResNet-101 & Cosine Margin & Influence loss on  ${\cal T}_{\rm old}$     \\
$\phi_{\rm old}^{\rm S}$  & - & 100\%  & 128 & ResNet-101 & SoftMax & -   \\
$\phi_{\rm new-\beta}^{\rm Cos-S}$ &$\phi_{\rm old}^{\rm S}$  & 100\%  & 128 & ResNet-101 & Cosine Margin & Influence loss on  ${\cal T}_{\rm old}$     \\
\bottomrule
\end{tabular}
\caption{`$\phi_{old}$': the compatible target model for New model.  
`$\phi_{{\rm new}-\beta}^{\rm R152}$': using ResNet-152 as backbone with the proposed BCT. 
`$\phi_{{\rm new}-\beta}^{\rm R152+256D}$': using ResNet-152 as backbone and feature dimension of 256 with the proposed BCT. 
`$\phi_{{\rm new}-\beta}^{\rm ReLU}$': adding a ReLU module after the embedding output of the new model when training with BCT.
`$\phi_{\rm old}^{\rm NS}$': the old model with normalized SoftMax classifier~\cite{wang2017normface}. 
`$\phi_{{\rm new}-\beta}^{\rm Cos-NS}$': the new model with cosine margin classifier~\cite{wang2018cosface} and trained by BCT with $\phi_{\rm old}^{\rm NS}$ as the old model.
`$\phi_{\rm old}^{\rm S}$': using standard softmax loss as training loss. 
`$\phi_{\rm new-\beta}^{\rm Cos-S}$': the new model with cosine margin classifier class~\cite{wang2018cosface} and BCT with $\phi_{\rm old}^{\rm S}$ as the old model.
}
\label{tab:bct_robust_setting}
\end{subtable}
\hfill
\\
\begin{subtable}[t]{0.5\linewidth} {\hspace{0pt}}
\small
\setlength{\tabcolsep}{1.pt}
\begin{tabular}{lcccc}
\toprule
\multirow{2}{*}{\shortstack{Comparison Pair}} 
& \multirow{2}{*}{\shortstack{Veri.\\ Acc. }} 
&\multirow{2}{*}{\shortstack{Backward\\Compatible?}}
& \multirow{2}{*}{\shortstack{Update\\Gain (\%) }}  
& \multirow{2}{*}{\shortstack{Absolute\\Gain}} \\
& \multicolumn{1}{c}{ }  \\
\midrule
$(\phi_{\rm old}, \phi_{\rm old})$ (Lower B.)&77.86 & - & - & - \\
$(\phi_{{\rm new}-\beta}^{\rm R152}, \phi_{\rm old})$ & 80.54 & $\surd$ & 29.45 & 2.68\\
$(\phi_{{\rm new}-\beta}^{\rm R152+256D}, \phi_{\rm old})$ & 80.92 & $\surd$ & 33.63 & 3.06 \\
$(\phi_{{\rm new}-\beta}^{\rm ReLU}, \phi_{\rm old})$  & 34.70 & $\times$ & - & -\\
\hline
$(\phi_{\rm old}^{\rm NS}, \phi_{\rm old}^{\rm NS}) $  (Lower B.) & 80.10 & - & - &  -\\
$(\phi_{\rm new-\beta}^{\rm Cos-NS}, \phi_{\rm old}^{\rm NS})$ & 81.81 & $\surd$ & 24.93 & 1.71 \\
\hline
($\phi_{\rm old}^{\rm S}$ ,$\phi_{\rm old}^{\rm S}$)  (Lower B.) & 73.27& - & - & - \\
($\phi_{\rm new-\beta}^{\rm Cos-S}$, $\phi_{\rm old}^{\rm S}$)  &  67.11 & $\times$ & - & -   \\
\hline
$(\phi_{\rm new}^*, \phi_{\rm new}^*)$ \begin{footnotesize}(Upper B.)\end{footnotesize} &86.96 & -& - & 9.1\\
\bottomrule
\end{tabular}
\caption{Experiments on the IJB-C 1:1 verification task. The verification accuracy evaluation metric is TAR (\%)@FAR=$10^{-4}$.
}
\label{tab:bct_robust_result_1v1}
\end{subtable}
\begin{subtable}[t]{0.5\linewidth}
\setlength{\tabcolsep}{1.pt}
\begin{tabular}{lcccc}
\toprule
\multirow{2}{*}{\shortstack{Comparison Pair}} 
& \multirow{2}{*}{\shortstack{Search\\Acc.}}

&\multirow{2}{*}{\shortstack{Backward\\Compatible?}} 
& \multirow{2}{*}{\shortstack{Update\\Gain (\%)}} 
& \multirow{2}{*}{\shortstack{Absolute\\Gain}}\\
& \multicolumn{1}{c}{ }  \\
\midrule
$(\phi_{\rm old}, \phi_{\rm old})$  (Lower B.) &59.34 & - & - & - \\
$(\phi_{{\rm new}-\beta}^{\rm R152}, \phi_{\rm old})$ & 68.71 & $\surd$ & 53.42 & 9.37 \\
$(\phi_{{\rm new}-\beta}^{\rm R152+256D}, \phi_{\rm old})$ & 69.45 & $\surd$ & 57.63 & 10.11 \\
$(\phi_{{\rm new}-\beta}^{\rm ReLU}, \phi_{\rm old})$ & 17.73 & $\times$ & - & -\\
\hline
($\phi_{\rm old}^{\rm NS}$ ,$\phi_{\rm old}^{\rm NS}$)  (Lower B.)  &64.32  & -  &- & -\\
$(\phi_{\rm new-\beta}^{\rm Cos-NS}, \phi_{\rm old}^{\rm NS})$ & 71.16  & $\surd$&54.46 & 6.84\\
\hline
($\phi_{\rm old}^{\rm S}$ ,$\phi_{\rm old}^{\rm S}$)  (Lower B.) & 54.16 & - & - & -\\
($\phi_{\rm new-\beta}^{\rm Cos-S}$, $\phi_{\rm old}^{\rm S}$)   &  45.46 & $\times$ & - & - \\
\hline
$(\phi_{\rm new}^*, \phi_{\rm new}^*)$ \begin{footnotesize}(Upper B.)\end{footnotesize} &76.88 & - & - & 17.54 \\
\bottomrule
\end{tabular}
\caption{Experiments on the IJB-C 1:N retrieval task. The search accuracy evaluation metric is TPIR(\%)@FPIR=$10^{-2}$.}
\label{tab:bct_robust_result_1vn}
\end{subtable}
\vspace{-10pt}
\caption{Robustness analysis of BCT against different training factors. We train new models with BCT while changing the network structure, feature dimensions, data amount and supervision loss, respectively. The training details for different models are listed in Tab.~\ref{tab:bct_robust_setting}. 
}
\label{tab:bct_robust}
\vspace{-15pt}
\end{table*}

\subsection{Learning with BCT}
\label{sec:exp_proposed}
We now experiment with the proposed BCT framework for backward compatible representation learning, starting from its basic form described in Sect.~\ref{sec:bct_basic}.
We use the same old model in the previous section. For the new model, 
we train it with the objective function described in Eq.~\eqref{eq:bct_loss}.  This model is called $\phi_{{\rm new}-\beta}$.
As illustrated in Tab.~\ref{tab:bct_baselines_result_1v1} and Tab.~\ref{tab:bct_baselines_result_1vn}, the model pair, $(\phi_{{\rm new}-\beta}, \phi_{\rm old})$, satisfies the backward compatibility criterion in Eq.~\eqref{eq:compatible_set}. 
Additionally, we observe update gains of $26.26\%$ and $44.98\%$ on the 1:1 verification and 1:N search protocols respectively.

We also evaluate a baseline approach adapted from~\cite{li2017learning}. The model trained with this approach is denoted as $\phi_{{\rm new-LwF}}$. It uses fixed $\phi_{\rm old}$ and its classifier $w_{c \ \rm old}$ to output soft labels of newly added samples $x \in \cal{T_{\rm new}} \setminus \cal{T_{\rm old}}$ as pseudo labels for training $\phi_{{\rm new-LwF}}$. 

From Tab.\ref{tab:bct_baselines_result_1v1} and \ref{tab:bct_baselines_result_1vn} we can see that model pair $(\phi_{{\rm new-LwF}}, \phi_{\rm old})$ does not satisfy the empirical backward compatibility criterion. Showing that directly adapting methods from the continual learning task does not work out of the box. However, it is able to improves to some extend the backward comparison accuracy, suggesting that the knowledge distillation used in continual learning could be useful in BCT.
We further investigate its application in the following experiments.

\textbf{BCT with newly added training data.}
In Sect.~\ref{sec:bct_dataset} we described two instantiations of BCT which can work with the new classes in the growing \embtraining. 
The first is using the synthesised classifier, we name the new model trained with this form of BCT as $\phi_{{\rm new}-\beta-sys}$. 
The second applies the idea of knowledge distillation to bypass obtaining classifier parameters for the new classes in the \embtraining. 
We name the new model trained with this form of BCT as $\phi_{{\rm new}-\beta-kd}$. 
The backward compatibility test results are summarized in Tab.~\ref{tab:bct_baselines}.
We can see that both new models can achieve backward compatibility. 
By fully utilizing the additional training data, they also lead to higher update gain (30.00$\%$ for $\phi_{{\rm new}-\beta-sys}$ and 27.25$\%$ for $\phi_{{\rm new}-\beta-kd}$) compared with the basic form of BCT (26.26$\%$ for $\phi_{{\rm new}-\beta}$).

\textbf{Does BCT hurt the accuracy of new models?}\label{sec:upper_bound}
One natural question is whether the influence loss is detrimental to the new model's recognition performance. 
We assess this by performing standard face recognition experiments on the 1:1 and 1:N protocols by extracting embedding only using the new models. 
This process can be considered as performing the backfill, or the paragon setting as described in Sect.~\ref{sec:baseline}.
The results are summarized in Tab.~\ref{tab:bct_upper_bound_res}.
We can see that training without BCT still yields the best accuracy in this setting. So, we name the model trained without BCT as $\phi_{\rm new}^*$, to indicate that it is the paragon that achieves best accuracy among all variants of new models.
Note that models trained with the basic form of BCT, $\phi_{{\rm new}-\beta}$, only leads to less than $3\%$ drop of accuracy in both tasks.
The new models $\phi_{{\rm new}-\beta-sys}$ and $\phi_{{\rm new}-\beta-kd}$ further reduce the gap.

\begin{table}
\small
\renewcommand\arraystretch{0.8}
\begin{subtable}[t]{0.5\textwidth}{\hspace{0pt}}
\centering
\setlength{\tabcolsep}{3pt}
\begin{tabular}{lccc}
\toprule
New Model  &Old Model & Data  & Additional Loss  \\
\midrule
$\phi_{1}$ & -& 25\%  &  -   \\
$\phi_{2}$  & $\phi_{1}$& 50\%  & Influence loss on $ {\cal T}_1$   \\
$\phi_{3}$ & $\phi_{2}$ & 100\%  & Influence loss on $ {\cal T}_2$  \\
\bottomrule
\end{tabular}
\caption{ `$\phi_{1}$': the first version model trained with 25\% of training data. `$\phi_{2}$':  the second version model trained on  50\% of training data with BCT towards $\phi_{1}$.  `$\phi_{3}$':  the third version model trained on all training data with BCT towards $\phi_{2}$.  }
\label{tab:bct_chain_setting}
\end{subtable}
\hfill
\\
\begin{subtable}[t]{1.0\linewidth} {\hspace{10pt}}
\setlength{\tabcolsep}{2.0pt}
\begin{tabular}{lcccc}
\toprule

\multirow{2}{*}{\shortstack{Comparison Pair}} 
& \multirow{2}{*}{\shortstack{Veri.\\ Acc. }} 

&\multirow{2}{*}{\shortstack{Backward\\Compatible?}}
& \multirow{2}{*}{\shortstack{Update\\Gain (\%) }}  
& \multirow{2}{*}{\shortstack{Absolute\\Gain}} \\
& \multicolumn{1}{c}{ }  \\
\midrule
$(\phi_{1}, \phi_{1})$ &41.45 & - & -  & - \\

$(\phi_{2}, \phi_{1})$ &56.34&$\surd$ & 40.90 & 14.89\\
$(\phi_{3}, \phi_{1})$ & 53.98& $\surd$ &34.41 & 12.53\\
\hline
$(\phi_{2}, \phi_{2})$ & 75.96  & - & - & -  \\ 
$(\phi_{3}, \phi_{2})$ &80.40& $\surd$ &40.36 & 4.44\\ 
\hline
$(\phi_{2}^*, \phi_{2}^*)$ & 77.86  & - & -  & 36.41 \\ 
$(\phi_{3}^*, \phi_{3}^*)$ &86.96& - & - & 9.1\\ 
\bottomrule
\end{tabular}
\caption{Experiments on the IJB-C 1:1 verification task for chain update ability test.}
\label{tab:bct_chain_result_1v1}
\end{subtable}
\begin{subtable}[t]{1.0\linewidth} {\hspace{10pt}}
\setlength{\tabcolsep}{2.0pt}
\begin{tabular}{lcccc}
\toprule
\multirow{2}{*}{\shortstack{Comparison Pair}} 
& \multirow{2}{*}{\shortstack{Search\\Acc.}}

&\multirow{2}{*}{\shortstack{Backward\\Compatible?}} 
& \multirow{2}{*}{\shortstack{Update\\Gain (\%)}} 
& \multirow{2}{*}{\shortstack{Absolute\\Gain}}\\
& \multicolumn{1}{c}{ }  \\

\midrule
$(\phi_{1}, \phi_{1})$ &22.57 & - & - & -  \\
$(\phi_{2}, \phi_{1})$ &39.00 & $\surd$ & 44.68 & 16.43\\
$(\phi_{3}, \phi_{1})$ &36.10 & $\surd$ & 36.80 & 13.53 \\
\hline
$(\phi_{2}, \phi_{2})$ & 56.07 & - & -  & -   \\
$(\phi_{3}, \phi_{2})$ &66.09 & $\surd$ & 48.15 & 10.02\\ 
\hline
$(\phi_{2}^*, \phi_{2}^*)$ & 59.34  & - & -  & 36.77  \\ 
$(\phi_{3}^*, \phi_{3}^*)$ &76.88& - &- & 17.54 \\ 
\bottomrule
\end{tabular}
\caption{Experiments on the IJB-C 1:N verification task for chain update ability test. 
}
\label{tab:bct_chain_result_1vn}
\end{subtable}
\vspace{-10pt}
\caption{ Experiments on multi-model compatibility between three models trained with growing amount of data.  In Tab.~\ref{tab:bct_chain_setting}, we illustrate the training details of all the models we trained. In Tab.~\ref{tab:bct_chain_result_1v1} and Tab.~\ref{tab:bct_chain_result_1vn}, we verify the compatibility of the trained models. Please be noted that the update gains are calculated by comparing ($\phi_{1}, \phi_{2}$), ($\phi_{1}, \phi_{3}$), and ($\phi_{2}, \phi_{3}$). }
\label{tab:bct_chain}
\vspace{-15pt}
\end{table}

\subsection{Extensions of BCT}
In the following experiments we explore whether BCT can be applied to different types of model training and achieve {\em multi-model} compatibility.

\textbf{Other changes in training $\phi_{\rm new}$.}
Besides increasing the size of the \embtraining, the new model $\phi_{\rm new}$ could have a new model architecture ({\em e.g.}, depth), a different loss function for supervision, or different embedding dimensions.  
We experiment the effect of these factors on BCT. 
For network architectures, we test a new model using ResNet-152~\cite{he2016deep} instead of ResNet-101~\cite{he2016deep} in previous experiments, denoted as $\phi_{{\rm new}-\beta}^{\rm R152}$.
In terms of loss types, we test using Norm-Softmax Loss~\cite{wang2017normface} for the old model, $\phi_{\rm old}^{\rm NS}$, and Cosine Margin Loss~\cite{wang2018cosface} for the new one, $\phi_{{\rm new}-\beta}^{\rm Cos}$.
In terms of embedding dimension, we test increasing the dimensions from 128 to 256 in the new model. Note that when the new model's feature dimension is changed, we will not be able to directly feed it to $w_{c \ \rm old}$. Here, we simply take the first 128 elements of the new model's feature to feed into $w_{c \ \rm old}$ during backward-compatible training and testing. We tried another approach of adding a linear transformer in training to match the feature dimension but without success. 

We also test the cases of changing several factors together, denoted as $\phi_{{\rm new}-\beta}^{\rm R152+256D}$.
The results are shown in Tab.~\ref{tab:bct_robust}. BCT can make most of new models backward compatible, even when several factors change simultaneously. 
This shows that BCT can be used as a general framework for achieving backward compatible representation learning. 

There are two failure cases where backward compatibility is not achieved in the pairs: 1)($\phi_{\rm new-\beta}^{\rm Cos-S}$, $\phi_{\rm old}^{\rm S}$), which uses Softmax loss~\cite{krizhevsky2012imagenet} for the old model and Cosine Margin Loss~\cite{wang2018cosface} for new model; this is possibly due to the drastic change of form of the loss functions. 2) $(\phi_{{\rm new}-\beta}^{\rm ReLU}, \phi_{\rm old})$, which adds ReLU activation on the embedding of the new model. The latter is possibly due to the distributional shift introduced by the ReLU activation in the new model, which makes it difficult for the new model with non-negative embedding vector elements to be compatible with the old model. This suggests that additional work is needed to expand the set of models that BCT can support.

\begin{table}[t!]
\centering
\footnotesize
\setlength{\tabcolsep}{1.pt}
\begin{tabular}{lccccc}
\toprule
\multirow{2}{*}{\shortstack{Comparison Pair}}   & \multirow{2}{*}{\shortstack{Mean  AP (\%)}} & \multirow{2}{*}{\shortstack{Backward \\ Compatible?}}  & 
\multirow{2}{*}{\shortstack{Update \\ Gain (\%)}} & \multirow{2}{*}{\shortstack{Absolute \\ Gain}}  & \multicolumn{1}{c}{ }  \\\\
\midrule
$(\phi_{\rm old}, \phi_{\rm old})$ & 42.9  & - & - & -   \\ \hline
$(\phi_{\rm old}, \phi_{\rm new}^*)$ &26.7 & $\times$ & - & -  \\
$(\phi_{\rm old}, \phi_{\rm new-\beta})$ & 45.0 & $\surd$ &  12.0 & 2.1\\ \hline
$(\phi_{\rm new-\beta}, \phi_{\rm new-\beta})$ &60.1 & -&  - & - \\ 
$(\phi_{\rm new}^*, \phi_{\rm new}^*)$ &60.3 & -& - & -\\
\bottomrule
\end{tabular}
\vspace{-5pt}
\caption{Backward compatibility tests on Market-1501 person re-identification dataset~\cite{zheng2015scalable}. Models $\phi_{old}$, $\phi_{new}^*$  are trained following~\cite{xiao2017joint} with $50\%$ and $100\%$ of training data, without BCT. $\phi_{new-\beta}$ is trained with $100\%$ of the training data and with BCT. Person search mean average precision (mean AP) is the accuracy metric. }
\label{tab:reid_market1501}
\vspace{-15pt}
\end{table}

\begin{figure}[t!]
\centering
\begin{tabular}{c@{\hspace{0mm}}c}

   &\includegraphics[scale=0.4]{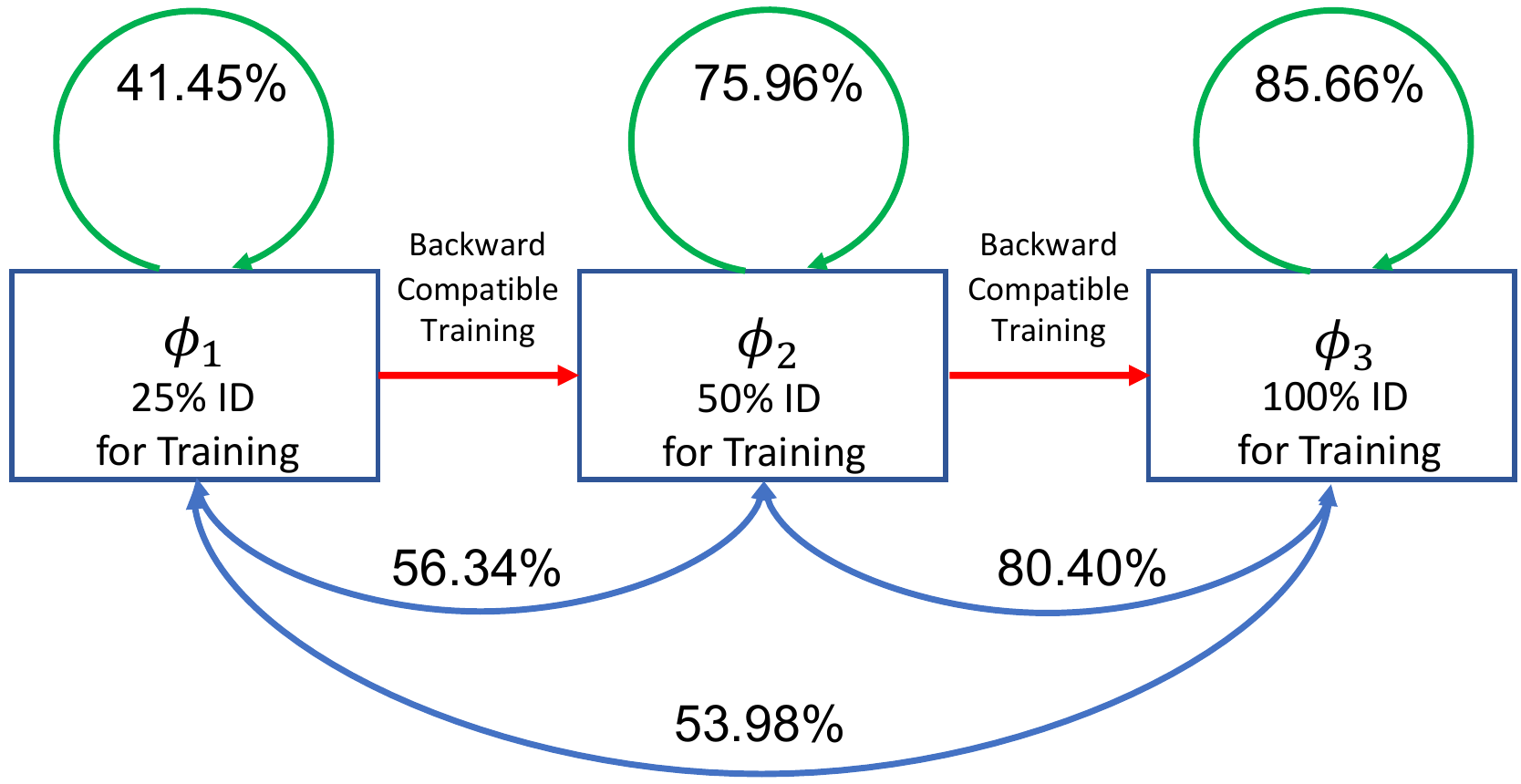}\\

\end{tabular}
\vspace{-15pt}
   \caption{Visualization of the multi-model compatibility experiment results. 
   We trained three models $\phi_{1}$, $\phi_{2}$, and $\phi_{3}$.
   The backward compatible training is enforced as shown above. 
   On the blue arrows we mark the backward compatibility test accuracy on the IJB-C 1:1 face verification benchmark, the green arrows mark the accuracy of the paragon settings.}
\label{fig:chain}
\vspace{-15pt}
\end{figure}

\textbf{Towards multi-model and sequential compatibility.}
Here we investigate a simple case of three model versions. 
The first version $\phi_{1}$ is trained with ${\cal T}_1$, which is a randomly sampled  $25\%$ subset of the IMDBFace dataset~\cite{wang2018devil}. 
The second version $\phi_{2}$ is trained  with ${\cal T}_2$, a $50\%$ subset. 
And the third version $\phi_{3}$ is trained with ${\cal T}_3$, which is the full IMDBFace dataset~\cite{wang2018devil}. 
We train $\phi_{2}$ using BCT with $\phi_{1}$ and train $\phi_{3}$ using BCT with $\phi_{2}$. 
Thus, in this process $\phi_{3}$ has no direct influence from $\phi_{1}$.
The backward compatibility test results are shown in Tab.~\ref{tab:bct_chain} and Fig.~\ref{fig:chain}. 
We observe that, by training with BCT, the last model $\phi_{3}$ is transitively compatible with $\phi_{1}$ even though $\phi_{1}$ is not directly involved in training $\phi_{3}$.
It shows that transitive compatibility between multiple models is indeed achievable through BCT, which could enable sequential update of the embedding models.

\textbf{BCT in other open-set recognition tasks.}
We validate the BCT method on the person re-identification task using the Market-1501~\cite{zheng2015scalable} benchmark. We train an old embedding model following~\cite{xiao2017joint} with $50\%$ of training data and two new embedding models with $100\%$ of the new training data. Search mean average precision (mean AP) is used as the accuracy metric. Table~\ref{tab:reid_market1501} shows the results of backward compatibility test.
We observe that the $\phi_{new-\beta}$ trained with BCT achieves backward compatibility without sacrificing its own search accuracy. 
This suggests that BCT can be a general approach for open-set recognition problems.

\vspace{-2pt}
\section{Discussion}
\vspace{-1pt}

We have presented a method for achieving backward-compatible representation learning, illustrated specific instances, and compared them with both baselines and paragons. Our approach has several limitations. The first is the accuracy gap of the new models trained with BCT relative to the new model oblivious of previous constraints.  Though the gap is reduced by slightly more sophisticated forms of BCT, there is still work wo to be done in characterizing and achieving the attainable accuracy limits.

\newpage
{\small
\bibliographystyle{ieee_fullname}
\bibliography{egbib}
}

\clearpage
\begin{appendices}

\appendix   
\appendixpage
\addappheadtotoc

\section{Implementation Details}

In Section 2.5, we describe how to use the influence loss on newly added training data for BCT to 1) compute synthesized classifier weights with the old model, and 2) use knowledge distillation.

When we have a newly added training example whose class is not in the old embedding training set, we feed the new image into the old model $\phi_{\rm old}$ and old classifier $w_{{\rm c \ old}}$ to obtain the classifier responses. Then, we can provide supervision signal to the new model $\phi_{\rm new}$ for this image by feeding the new model's embedding to the old classifier $w_{{\rm c \ old}}$ and compute the knowledge distillation loss (cross-entropy with temperature-modulated SoftMax) between the two response vectors as the influence loss.
Because we are using the cosine margin loss~\cite{wang2018cosface} in the experiments which also has a temprature parameter, we set the temperature parameter in knowledge distillation to the same as the one in the cosine margin loss, which is $32$.

\section{Partial Backfilling}
\begin{figure}[t]
\centering
\begin{tabular}{c}
  \includegraphics[width=1\linewidth]{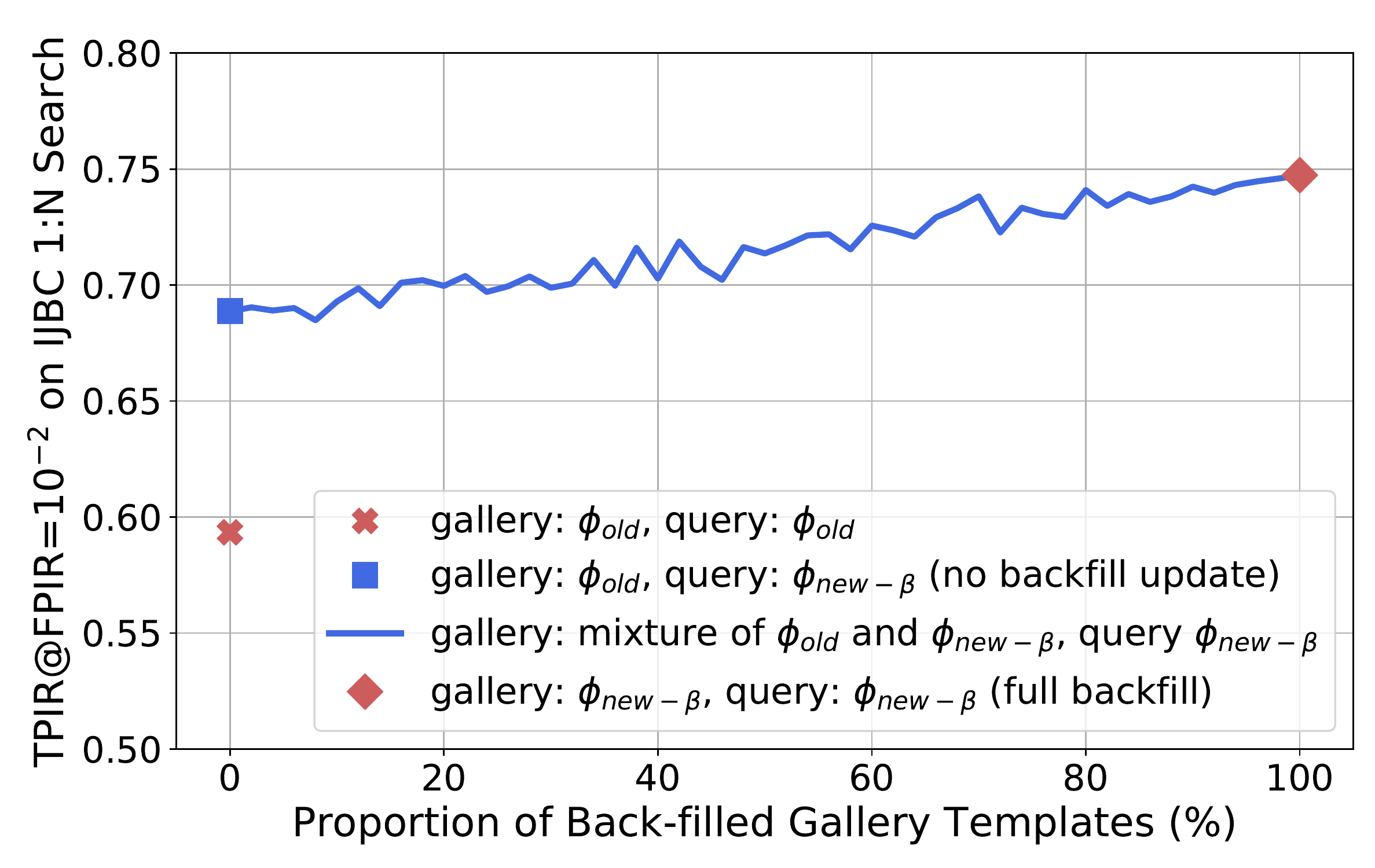}\\
\end{tabular}
   \caption{
   The curve for face search accuracy vs. backfill proportions. 
   We gradually backfill the gallery set from $0\%$ classes (same as the backward compatibility test or no-backfill update) to $100\%$ (fully backfill or the paragon setting). 
   Face search accuracy is measured between every $2\%$ of partial backfill. 
   The red cross shows the face search accuracy of the old model. 
   The red diamonds marks the face search accuracy of the new embedding model.
   }
\label{fig:partial_backfill_curve}
\end{figure}

Partial backfilling happens when only a part of the gallery classes have been processed by the new model $\phi_{new}$. We test whether queries from the backward compatible new model $\phi_{new-\beta}$ can work with partially backfilled gallery sets. In Fig.~\ref{fig:partial_backfill_curve}, we illustrate the search accuracy on gallery sets of different backfill ratios.  As higher percentages of the gallery set are backfilled, search accuracy grow proportionally  towards the accuracy of the fully backfilled case. This suggests that one can upgrade to a new model, immediately benefiting from the improved accuracy, and optionally backfill the old gallery gradually in the background until paragon performance is achieved.

\section{Detailed Benchmark Results}

Due to space limitations, in the main paper we only report one specific operating point for each metric in the evaluation, \emph{e.g.}, ${\rm TAR}@ {\rm FAR}=10^{-4}$ for face verification and ${\rm TPIR}@ {\rm FPIR}=10^{-2}$ for face identification. 
Here we report results at additional operating points on the IJB-C~\cite{maze2018iarpa} benchmark. 
In Table ~\ref{tab:bct_baselines_append}, we show the performance of different compared baselines and our proposed method. 
In Table ~\ref{tab:bct_abstudy_append} and Table ~\ref{tab:bct_chain_append}, we show the performance of extensions of our proposed backward-compatible trainting process. In Table ~\ref{tab:bct_abstudy_append}, we illustrate extensions of BCT to different model depths, feature dimensions and supervision losses. In Table ~\ref{tab:bct_chain_append}, we show extensions to multi-model compatibility towards sequential updating.

\begin{table*}[t]
\begin{center}
\begin{tabular}{lccc}
\hline
 & Continual Learning& Domain Adaptation & BCRL   \\
\hline
Access to all old model parameters & Yes & Yes & Not required \\
Access to old training data & Not required & Yes & Yes \\
Re-processing of test data? & Yes & Yes & Not required\\
Consistent output & Yes & Not required & Yes \\
Compatible representation & Not required & Not required & Yes \\
\hline
\end{tabular}
\end{center}
\caption{The differences between backward compatible representation learning (BCRL) , continual Learning, and domain adaptation. In this table, we list several features of the task each problem deals with to illustrate the difference.}
\label{tab:prob_diff}
\vspace{0pt}
\end{table*}


\begin{table*}
\small
\setlength{\tabcolsep}{3pt}
\begin{subtable}{1.0\textwidth}\centering
\begin{tabular}{llccccc}
\toprule
New Model  &Old Model & Training Data Usage & Feat. Dim. & Model Arc. & Classifier & Additional Loss \\
\midrule
$\phi_{\rm old}$ & -& 50\% & 128  & ResNet-101 & Cosine Margin &  - \\
$\phi_{{\rm new}-\beta}^{\rm R152} $ &$\phi_{\rm old}$& 100\%  & 128 & ResNet-152 & Cosine Margin   & Influence loss on  ${\cal T}_{\rm old}$   \\
$\phi_{{\rm new}-\beta}^{\rm R152+256D}$  &$\phi_{\rm old}$ & 100\%  & 256 & ResNet-152 & Cosine Margin   & Influence loss on  ${\cal T}_{\rm old}$   \\
$\phi_{{\rm new}-\beta}^{\rm ReLU}$  &$\phi_{\rm old}$ & 100\%  & 128 & ResNet-152 & Cosine Margin   &  Influence loss on  ${\cal T}_{\rm old}$    \\
\hline
$\phi_{\rm old}^{\rm NS}$  & - & 100\%  & 128 & ResNet-101 & Norm-Softmax  &  -   \\
$\phi_{{\rm new}-\beta}^{\rm Cos-NS}$& $\phi_{\rm old}^{\rm NS}$   & 100\%  & 128 & ResNet-101 & Cosine Margin & Influence loss on  ${\cal T}_{\rm old}$     \\
$\phi_{\rm old}^{\rm S}$  & - & 100\%  & 128 & ResNet-101 & SoftMax & -   \\
$\phi_{\rm new-\beta}^{\rm Cos-S}$ &$\phi_{\rm old}^{\rm S}$  & 100\%  & 128 & ResNet-101 & Cosine Margin & Influence loss on  ${\cal T}_{\rm old}$     \\
\hline
$\phi_1^{\rm \ \text{25\%}}$ & - &  25\% & 128  & ResNet-101 & Cosine Margin &  -   \\
$\phi_{{\rm new-\beta}}^\text{25\%}$ & $\phi_1^{\rm \ \text{25\%}}$& 100\% & 128  & ResNet-101 & Cosine Margin & Influence loss on  ${\cal T}_{\rm old \ 25\%}$    \\
$\phi_1^{\rm \ \text{90\%}}$  & - & 90\% & 128  & ResNet-101 & Cosine Margin  &  -  \\
$\phi_{{\rm new-\beta}}^\text{90\%}$  & $\phi_1^{\rm \ \text{90\%}}$& 100\% & 128  & ResNet-101 & Cosine Margin &  Influence loss on  ${\cal T}_{\rm old \ 90\%}$   \\
\bottomrule
\end{tabular}
\caption{`Old Model': the compatible target model for New model.  
`$\phi_{{\rm new}-\beta}^{\rm R152}$': using ResNet-152 as backbone with the proposed BCT. 
`$\phi_{{\rm new}-\beta}^{\rm R152+256D}$': using ResNet-152 as backbone and feature dimension of 256 with the proposed BCT. 
`$\phi_{{\rm new}-\beta}^{\rm ReLU}$': adding a ReLU module after the embedding output of the new model when training with BCT.
`$\phi_{\rm old}^{\rm NS}$': the old model with normalized SoftMax classifier~\cite{wang2017normface}. 
`$\phi_{{\rm new}-\beta}^{\rm Cos-NS}$': the new model with cosine margin classifier~\cite{wang2018cosface} and trained by BCT with $\phi_{\rm old}^{\rm NS}$ as the old model.
`$\phi_{\rm old}^{\rm S}$': using standard softmax loss as training loss. 
`$\phi_{\rm new-\beta}^{\rm Cos-S}$': the new model with cosine margin classifier class~\cite{wang2018cosface} and BCT with $\phi_{\rm old}^{\rm S}$ as the old model.
`$\phi_1^{\rm \ \text{25\%}}$': Model trained by 25\% of training data. 
`$\phi_1^{\rm \ \text{90\%}}$ ':  Model trained by 90\% of training data.
}
\label{tab:bct_abstudy_setting_append}
\end{subtable}
\hfill
\\ 
\begin{subtable}[t]{0.5\linewidth}
\centering
\begin{tabular}{lcccc}
\toprule
\multirow{3}{*}{Comparison Pair} & \multicolumn{4}{c}{IJB-C 1:1 Verification}  \\
&\multicolumn{4}{c}{ TAR (\%) @ FAR=} \\ \cmidrule(r){2-5} 
 &$10^{-5}$ & $10^{-4}$ & $10^{-3}$ & $10^{-2}$ \\ 
\midrule
$(\phi_{\rm old}, \phi_{\rm old})$ &59.41&77.86 &88.80&95.35 \\
$(\phi_{{\rm new}-\beta}^{\rm R152}, \phi_{\rm old})$ &66.44 & 80.54 & 89.87 & 95.71\\
$(\phi_{{\rm new}-\beta}^{\rm R152+256D}, \phi_{\rm old})$ & 68.45 & 80.92&89.83&95.84\\
$(\phi_{{\rm new}-\beta}^{\rm ReLU}, \phi_{\rm old})$ & 17.02&34.70&59.82&83.69\\
\hline
($\phi_{\rm old}^{\rm S}$ ,$\phi_{\rm old}^{\rm S}$) & 57.30 & 73.27&85.91&94.45 \\
($\phi_{\rm new-\beta}^{\rm Cos-S}$, $\phi_{\rm old}^{\rm S}$) & 48.08& 67.11 & 84.01 & 94.39 \\
($\phi_{\rm old}^{\rm NS}$ ,$\phi_{\rm old}^{\rm NS}$)  & 65.56 & 80.10 & 89.80 & 95.61 \\
$(\phi_{\rm new-\beta}^{\rm Cos-NS}, \phi_{\rm old}^{\rm NS})$  & 69.72 & 81.81 & 90.39 & 96.10 \\
\hline
$(\phi_1^{\rm \ \text{25\%}}, \phi_1^{\rm \ \text{25\%}})$ & 29.40&51.77&73.30&88.97  \\
$(\phi_{{\rm new-\beta}}^\text{25\%}, \phi_1^{\rm \ \text{25\%}})$&33.51&52.21&72.78&88.84 \\
$(\phi_1^{\rm \ \text{90\%}}, \phi_1^{\rm \ \text{90\%}})$& 74.49& 86.58&93.24&97.03  \\
$(\phi_{{\rm new-\beta}}^\text{90\%}, \phi_1^{\rm \ \text{90\%}})$&74.61&86.27&93.19&97.08\\ 
\bottomrule
\end{tabular}
\caption{Experiments on the IJB-C 1:1 verification task. }
\label{tab:bct_abstudy_result_1v1_append}
\end{subtable}
\begin{subtable}[t]{0.5\linewidth}
\centering
\setlength{\tabcolsep}{4.pt}
\begin{tabular}{lccccc}
\toprule
 \multirow{3}{*}{Comparison Pair}& \multicolumn{5}{c}{IJB-C 1:N Retrieval}  \\
 &\multicolumn{3}{c}{TPIR (\%) @ FPIR=} & \multicolumn{2}{c}{Retrieval Rate (\%)} \\\cmidrule(r){2-4} \cmidrule(r){5-6} 
 &$10^{-3}$ & $10^{-2}$ & $10^{-1}$ & Rank-1 & Rank-5 \\ 
\midrule
$(\phi_{\rm old}, \phi_{\rm old})$ &36.90&59.34&79.18&87.25&92.83  \\
$(\phi_{{\rm new}-\beta}^{\rm R152}, \phi_{\rm old})$ & 57.22&68.71&82.58&89.19&94.10\\
$(\phi_{{\rm new}-\beta}^{\rm R152+256D}, \phi_{\rm old})$  &54.77& 69.45&83.60&89.12&94.07 \\
$(\phi_{{\rm new}-\beta}^{\rm ReLU}, \phi_{\rm old})$ & 6.98&17.73&33.13&52.58&73.32\\
\hline
($\phi_{\rm old}^{\rm S}$ ,$\phi_{\rm old}^{\rm S}$)  & 31.53 & 54.16& 73.88 & 85.42 & 92.69 \\
($\phi_{\rm new-\beta}^{\rm Cos-S}$, $\phi_{\rm old}^{\rm S}$)& 31.44 & 45.46 & 69.19 & 85.37 & 92.80\\
($\phi_{\rm old}^{\rm NS}$ ,$\phi_{\rm old}^{\rm NS}$) & 40.16&64.32&81.36&89.35&94.39 \\
$(\phi_{\rm new-\beta}^{\rm Cos-NS}, \phi_{\rm old}^{\rm NS})$  & 54.34 & 71.16 & 83.40 & 90.24&94.74 \\
\hline
$(\phi_1^{\rm \ \text{25\%}}, \phi_1^{\rm \ \text{25\%}})$ &11.34 & 26.84 & 54.00 & 71.67 & 82.94  \\
$(\phi_{{\rm new-\beta}}^\text{25\%}, \phi_1^{\rm \ \text{25\%}})$ &16.48&34.24&57.66&76.89&87.44 \\
$(\phi_1^{\rm \ \text{90\%}}, \phi_1^{\rm \ \text{90\%}})$& 57.81&74.52&87.28 &91.95 & 95.72  \\
$(\phi_{{\rm new-\beta}}^\text{90\%}, \phi_1^{\rm \ \text{90\%}})$&61.41&74.57&87.50&91.92&95.51\\ 
\bottomrule
\end{tabular}
\caption{Experiments on the IJB-C 1:N retrieval task.}
\label{tab:bct_abstudy_result_1vn_append}
\end{subtable}
\caption{Robustness analysis of our proposed method against different training factors. When we use the proposed Backward Compatible Training method to train the new model, we change the network structure, feature dimension, data amount and supervision loss, respectively. }
\label{tab:bct_abstudy_append}
\vspace{0pt}
\end{table*}

\begin{table}[t]
\small
\setlength{\tabcolsep}{3.5pt}
\begin{subtable}{1.0\linewidth}\hspace{0pt}
\centering
\renewcommand\arraystretch{0.8}
\setlength{\tabcolsep}{4.0pt}
\begin{tabular}{lccc}
\toprule
New Model & Old Model & Data & Additional Loss\\
\midrule
$\phi_{\rm old}$  & - & 50\%  & -\\
$\phi_{\rm new}^*$  & - & 100\%     & -\\
$\phi_{{\rm new}-\ell^2}$   &$\phi_{\rm old}$  & 100\%     & $\ell^2$ distance $\phi_{\rm old}$\\
$\phi_{{\rm new-LwF}}$ &$\phi_{\rm old}$ &  50\%  & Learning w/o Forgetting\\
$\phi_{{\rm new}-\beta}$ &$\phi_{\rm old}$  & 100\% &  Influence loss on ${\cal T}_{\rm old}$ \\
$\phi_{{\rm new}-\beta-kd}$ & $\phi_{\rm old}$ & 100\% &  Influence loss on ${\cal T}_{\rm new}$ \\
$\phi_{{\rm new}-\beta-sys}$ & $\phi_{\rm old}$ & 100\% &  Influence loss on ${\cal T}_{\rm new}$\\

\bottomrule
\end{tabular}
\vspace{0pt}
\caption{Training setting for different backward-compatible (new) models. 
`Old Model': The compatible target model for New model. 
`$\phi_{{\rm new}-\ell^2}$': The new model with cosine margin classifier~\cite{wang2018cosface} and regularized with $\ell^2$ distance to $\phi_{\rm old}$ output feature. 
`$\phi_{{\rm new-LwF}}$': The new model with cosine margin classifier~\cite{wang2018cosface} and adopt Learning w/o Forgetting~\cite{li2017learning} approach for the new model training.
`$\phi_{{\rm new}-\beta}$': The new model trained with proposed BCT.
`$\phi_{{\rm new}-\beta-kd}$': Trained with proposed BCT and use knowledge distillation to bypass obtaining soft supervision labels for the new classes in the new embedding training dataset. 
`$\phi_{{\rm new}-\beta-sys}$': Trained with proposed BCT and use the feature processed by $\phi_{\rm old}$ as the synthesised  classifier for new classes  in  the  growing  embedding  training dataset. 
}
\label{tab:bct_baselines_setting_append}
\end{subtable}
\hfill
\\
\begin{subtable}[t]{1.0\linewidth}
\centering
\setlength{\tabcolsep}{3pt}
\begin{tabular}{lcccc}
\toprule
\multirow{3}{*}{Comparison Pair} &  \multicolumn{4}{c}{IJB-C 1:1 Verification}  \\
 &\multicolumn{4}{c}{ TAR (\%) @ FAR=} \\ \cmidrule(r){2-5} 
 &$10^{-5}$ & $10^{-4}$ & $10^{-3}$ & $10^{-2}$ \\ 
\midrule
$(\phi_{\rm old}, \phi_{\rm old})$&59.41&77.86 &88.80&95.35 \\
\hline
$(\phi_{\rm new}^*, \phi_{\rm old})$&0.0&0.0&0.0&0.0  \\
$(\phi_{{\rm new}-\ell^2}, \phi_{\rm old})$ &0.5&3.10&10.32&31.98 \\
$(\phi_{{\rm new-LwF}}, \phi_{\rm old})$  & 57.09 & 77.26 & 88.65&95.46 \\
$(\phi_{{\rm new-\beta}}, \phi_{\rm old})$ (Proposed)&66.06&80.25&89.79&95.62  \\
$(\phi_{{\rm new}-\beta-kd}, \phi_{\rm old})$ (Proposed)&67.82&80.34&89.60&95.74\\
$(\phi_{{\rm new}-\beta-sys}, \phi_{\rm old})$ (Proposed)& 68.35&80.59&89.42&95.23\\
\hline
$(\phi_{\rm new}^*,\phi_{\rm new}^*)$&76.77&86.96&93.66&97.18  \\
\bottomrule
\end{tabular}
\caption{Experiments on the IJB-C 1:1 verification task. }
\label{tab:bct_baselines_result_1v1_append}
\end{subtable}
\centering
\begin{subtable}[t]{1.0\linewidth}
\setlength{\tabcolsep}{3pt}
\footnotesize
\begin{tabular}{lccccc}
\toprule
\multirow{3}{*}{Comparison Pair} & \multicolumn{5}{c}{IJB-C 1:N Retrieval}  \\
 &\multicolumn{3}{c}{TPIR (\%) @ FPIR=} & \multicolumn{2}{c}{Retrieval Rate (\%)} \\\cmidrule(r){2-4} \cmidrule(r){5-6} 
 &$10^{-3}$ & $10^{-2}$ & $10^{-1}$ & Rank-1 & Rank-5 \\ 
\midrule
$(\phi_{\rm old}, \phi_{\rm old})$&36.90&59.34&79.18&87.25&92.83  \\
\hline
$(\phi_{\rm new}^*, \phi_{\rm old})$&0.0&0.0&0.0 & 0.0& 0.01  \\
$(\phi_{{\rm new}-\ell^2}, \phi_{\rm old})$ &0.14&0.50&2.93&8.41&20.25  \\
$(\phi_{{\rm new-LwF}}, \phi_{\rm old})$  &35.89&59.27&79.00 &87.35&93.12\\
$(\phi_{{\rm new-\beta}}, \phi_{\rm old})$ (Proposed)&52.58&67.23&82.34&88.95&93.95\\
$(\phi_{{\rm new-\beta-kd}}, \phi_{\rm old})$ (Proposed)&56.20&69.02&82.50&89.01&93.96\\
$(\phi_{{\rm new-\beta-sys}}, \phi_{\rm old})$ (Proposed)&59.48&70.70&82.97&90.09&94.53\\
\hline
$(\phi_{\rm new}^*,\phi_{\rm new}^*)$&61.93&76.88&87.70&92.11&95.72  \\
\bottomrule
\end{tabular}
\caption{Experiments on the IJB-C 1:N search task.}
\label{tab:bct_baselines_result_1vn_append}
\end{subtable}
\caption{We experiment with different approaches towards compatibility of comparison pair. In (a), we illustrate the training details of all the models we trained. In (b), we show the benchmarking results of 1:1 verification on the IJB-C dataset~\cite{maze2018iarpa}. In (c), we show the benchmarking results of 1:N search on the IJB-C dataset.}
\label{tab:bct_baselines_append}

\end{table}


\begin{table}[t]
\centering
\begin{subtable}[t]{1.0\linewidth}
\setlength{\tabcolsep}{3pt}
\begin{tabular}{lccc}
\toprule
New Model  &Old Model & Data  & Additional Loss  \\
\midrule
$\phi_{1}$ & -& 25\%  &  -   \\
$\phi_{2}$  & $\phi_{1}$& 50\%  & Influence loss on $ {\cal T}_1$   \\
$\phi_{3}$ & $\phi_{2}$ & 100\%  & Influence loss on $ {\cal T}_2$  \\
\bottomrule
\end{tabular}
\label{tab:bct_chain_setting_append}
\caption{`Old Model': the compatible target model for New model.  `$\phi_{1}$': model trained with 25\% of training data. `$\phi_{2}$':  Model trained by  50\% of training data and proposed BCT with $\phi_{1}$.  `$\phi_{3}$':  Model trained by  all of training data and proposed BCT with $\phi_{2}$.  }
\end{subtable}
\hfill
\\
\begin{subtable}[t]{1.0\linewidth}
\setlength{\tabcolsep}{4pt}
\small
\centering
\begin{tabular}{llcccc}
\toprule
\multirow{3}{*}{Comparison Pair} & \multicolumn{4}{c}{IJB-C 1:1 Verification}  \\
&\multicolumn{4}{c}{ TAR (\%) @ FAR=} \\ \cmidrule(r){2-5} 
&$10^{-5}$ & $10^{-4}$ & $10^{-3}$ & $10^{-2}$ \\ 
\midrule
$(\phi_{1}, \phi_{1})$ & 29.24&41.45&73.40&89.06  \\
$(\phi_{2}, \phi_{1})$ & 36.13&56.34&75.14&89.54\\
$(\phi_{3}, \phi_{1})$ & 32.14&53.98&71.95&88.69\\ 
\hline
$(\phi_{2}, \phi_{2})$  &58.01&75.96&87.97&95.18\\
$(\phi_{3}, \phi_{2})$  &64.84&80.40&89.84&95.84 \\
$(\phi_{3}, \phi_{3})$  &73.61&85.66&92.88&96.90\\
\bottomrule
\end{tabular}
\caption{Experiments on the IJB-C 1:1 verification task.}
\label{tab:bct_chain_result_1v1_append}
\end{subtable}
\begin{subtable}[t]{1.0\linewidth}
\setlength{\tabcolsep}{3pt}
\small
\centering
\captionsetup{width=0.9\textwidth}
\begin{tabular}{lccccc}
\toprule
\multirow{3}{*}{Comparison Pair}& \multicolumn{5}{c}{IJB-C 1:N Retrieval}  \\
& \multicolumn{3}{c}{TPIR (\%) @ FPIR=} & \multicolumn{2}{c}{Retrieval Rate (\%)} \\\cmidrule(r){2-4} \cmidrule(r){5-6} 
& $10^{-3}$ & $10^{-2}$ & $10^{-1}$ & Rank-1 & Rank-5 \\ 
\midrule
$(\phi_{1}, \phi_{1})$ &11.30 &22.57&54.39&71.52&82.87  \\
$(\phi_{2}, \phi_{1})$ &19.21&39.00&59.85&78.22&87.81\\
$(\phi_{3}, \phi_{1})$ &14.85&36.10&56.35&77.18&87.48\\ 
\hline
$(\phi_{2}, \phi_{2})$  &38.59&56.07&78.50&86.97&92.49\\
$(\phi_{3}, \phi_{2})$  &50.81&66.09&82.97&89.26&94.11\\
$(\phi_{3}, \phi_{3})$  &59.22&74.12&86.70&91.61&95.53\\
\bottomrule
\end{tabular}
\caption{Experiments on the IJB-C 1:N verification task.}
\label{tab:bct_chain_result_1vn_append}
\end{subtable}
\caption{ Robustness analysis of our proposed method against different training factors. When we use the proposed Backward Compatible Training method to train the new model, we change the network structure, feature dimension, data amount and supervision loss, respectively. }
\label{tab:bct_chain_append}

\end{table}
\end{appendices}

\end{document}